\pdfoutput=1

\documentclass[11pt]{article}

\usepackage[final]{acl}

\usepackage{times}
\usepackage{latexsym}
\usepackage{graphicx}
\usepackage{array}
\usepackage{booktabs}
\usepackage{url}
\usepackage{xcolor}
\usepackage[many]{tcolorbox}

\usepackage[T1]{fontenc}

\usepackage[utf8]{inputenc}

\usepackage{microtype}

\usepackage{inconsolata}

\usepackage{graphicx}

\title{\textsc{HiligayNER}: A Baseline Named Entity Recognition Model for Hiligaynon}

\author{\textbf{James Ald Teves}\textsuperscript{\textbf{1}}, \textbf{Ray Daniel Cal}\textsuperscript{\textbf{1}}, \textbf{Josh Magdiel Villaluz}\textsuperscript{\textbf{1}}, \textbf{Jean Malolos}\textsuperscript{\textbf{2}}, \\ \textbf{Mico Magtira\textsuperscript{2}, Ramon Rodriguez\textsuperscript{2}, Mideth Abisado\textsuperscript{2} and Joseph Marvin Imperial\textsuperscript{2}} \\ 
  \textsuperscript{1}Silliman University \\
  \textsuperscript{2}National University Philippines \\
  \texttt{\url{jamesyteves@su.edu.ph}, \url{jrimperial@national-u.edu.ph}}
  }

\begin{document}
\maketitle
\begin{abstract}
The language of Hiligaynon, spoken predominantly by the people of Panay Island, Negros Occidental, and Soccsksargen in the Philippines, remains underrepresented in language processing research due to the absence of annotated corpora and baseline models. This study introduces \textbf{\textsc{HiligayNER}}, the first publicly available baseline model for the task of Named Entity Recognition (NER) in Hiligaynon. The dataset used to build \textsc{HiligayNER} contains over 8,000 annotated sentences collected from publicly available news articles, social media posts, and literary texts. Two Transformer-based models, mBERT and XLM-RoBERTa, were fine-tuned on this collected corpus to build versions of \textsc{HiligayNER}. Evaluation results show strong performance, with both models achieving over 80\% in precision, recall, and F1-score across entity types. Furthermore, cross-lingual evaluation with Cebuano and Tagalog demonstrates promising transferability, suggesting the broader applicability of \textsc{HiligayNER} for multilingual NLP in low-resource settings. This work aims to contribute to language technology development for underrepresented Philippine languages, specifically for Hiligaynon, and support future research in regional language processing.\footnote{Code and data: \url{https://github.com/jvlzloons/HiligayNER}}
\end{abstract}

\section{Introduction}
The coverage and representation of diverse regional languages play a key role in the widespread adoption of any AI-based technology across the globe. While English remains the most highly researched and high-resourced language, initiatives from the research community, such as the SEACrowd \cite{cahyawijaya-etal-2025-crowdsource,lovenia-etal-2024-seacrowd} for Southeast Asian languages, Masakhane \cite{adelani-etal-2023-masakhanews,adelani-etal-2021-masakhaner} for African languages, and Aya Project \cite{ustun-etal-2024-aya,singh-etal-2024-aya} for global participation, have effectively made its impact to close the AI language gap \cite{bassignana-etal-2025-ai,pava_et_al_2025}.


A recent survey of digital support levels of languages showed that regional Philippine languages are among the lowest representations worldwide \cite{r5}. One particular language is \textbf{Hiligaynon}\footnote{\url{https://www.ethnologue.com/language/hil/}}, which is an Austronesian regional language spoken by over 10 million people in Western Visayas, particularly Panay Island, Negros Occidental, and Soccsksargen \cite{r12,robles2012hiligaynon}. To initiate a step towards progress in Hiligaynon representation, researchers are encouraged to build resources and corpora for fundamental natural language processing tasks. One of these fundamental tasks is Named Entity Recognition (NER) or the task of automatic identification of textual mentions of persons, organizations, locations, and related categories \cite{nadeau2007survey,r21,yadav-bethard-2018-survey}. 


In this work, we present \textsc{HiligayNER}, the first publicly available NER corpus and finetuned models for Hiligaynon. Specifically, our contributions towards democratizing language resources for Hiligaynon are as follows:

\begin{enumerate}
    \item A compilation of cleaned sentence-level Hiligaynon dataset of over 8,000 entries from online publicly accessible news articles, social media posts, and translated texts.
    \item A compilation of span-level BIO-encoded annotations of the Hiligaynon dataset for the named entity recognition task (NER), specifically covering four entity categories (PER, ORG, LOC).
    \item Two finetuned multilingual Transformer-based models, mBERT and XLM-RoBERTa, for token-level sequence labeling of Hiligaynon texts.
\end{enumerate}

By releasing the dataset, model checkpoints, and evaluation scripts under an open license, we aim to supply the foundational tools required for broader NLP development in Western Visayas and the wider Philippine research community.

\section{Related Works}

Robust NER systems enable downstream applications such as knowledge-graph construction, information retrieval, and domain-specific analytics \cite{r7}. State-of-the-art performance is now achieved by combining lexicon-based gazetteers \cite{r8}, data-augmentation techniques \cite{r9}, and deep neural architectures ranging from BiLSTM-CRF \cite{r10} to multilingual transformer encoders \cite{r11,tan-etal-2024-cebbert}. Early Philippine NER studies concentrated almost exclusively on Tagalog, the national language. Statistical sequence models dominated. \cite{r13} applied Conditional Random Fields (CRF) to biographical texts and reported an F1 of 83\%, while \cite{r14} achieved 80.5\% with a maximum-entropy classifier on short-story data. Subsequent CRF experiments on a larger newswire corpus produced a lower but still respectable 75.7\% overall F1 \cite{r15}. 

Cebuano, the second most widely spoken native tongue in the country, received attention slightly later. Maynard’s rule-based adaptation of the ANNIE system yielded 69.1\% F1 on a modest test set \cite{r16}. Cross-lingual neural CRFs, transferring knowledge from Tagalog, pushed performance to 81.8\% \cite{r11}. More recently, \cite{r17} introduced a hybrid CNN–BiLSTM pipeline that surpassed 95\% precision and recall, albeit on only 200 manually annotated news articles. The largest Cebuano-based research to date is CebuaNER \cite{pilar-etal-2023-cebuaner}, which released a 4,258 article gold-standard corpus and baseline CRF/BiLSTM models that exceeded 70\% F1 across entity classes. These milestones underscore both the feasibility and the demand for regional-language NER resources in the Philippines.

In contrast, Hiligaynon still lacks a public NER corpus or baseline model. Computational work has been limited to tokenization heuristics and the compilation of morphosyntactic lexicons \cite{r12}; no peer-reviewed study has tackled entity annotation or sequence labelling. This shortfall hampers information-extraction pipelines for regional journalism, public administration, and social-media analytics in Western Visayas, where Hiligaynon is the dominant medium.

\begin{figure*}[!t]
    \centering
    \includegraphics[width=0.90\textwidth]{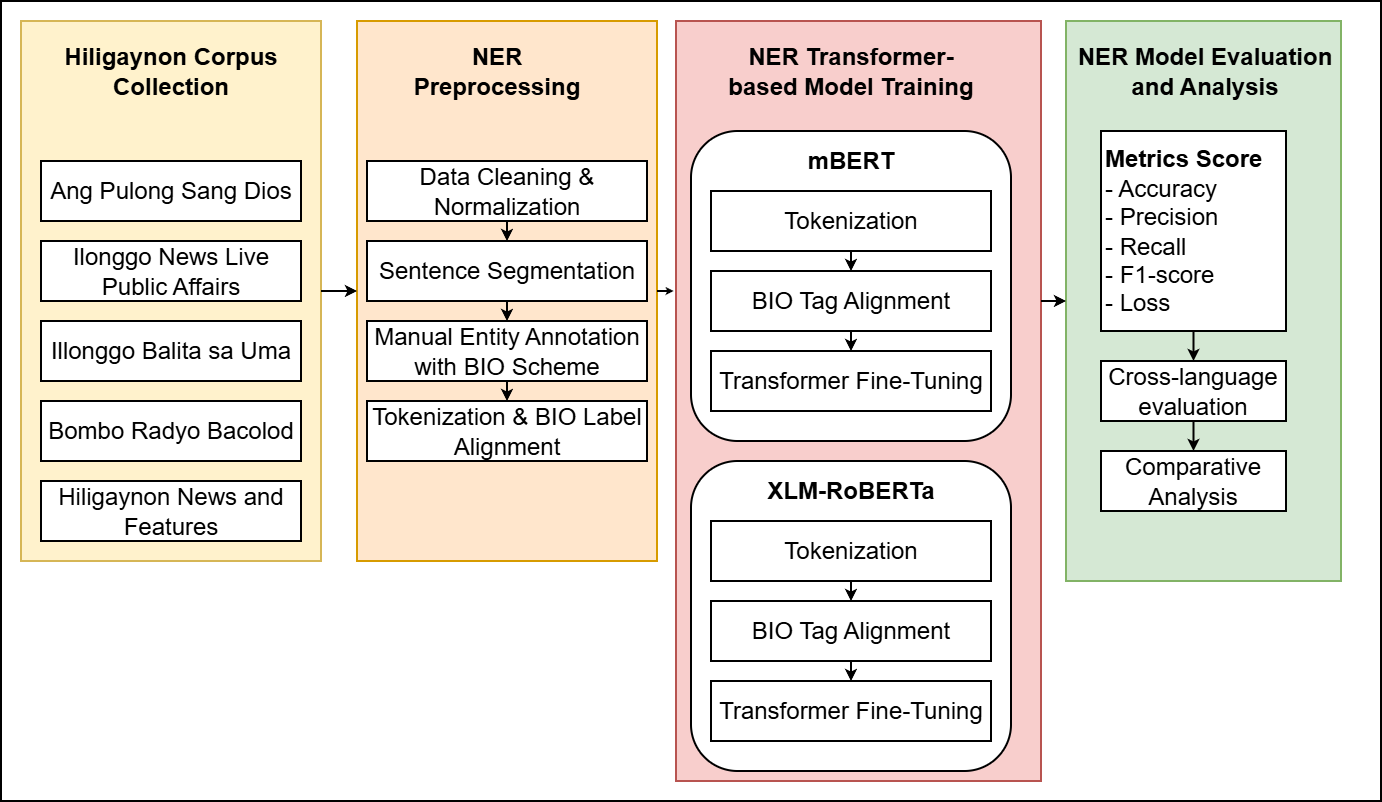}
    \caption{The overall methodology of developing \textsc{HiligayNER} using annotated news articles, social media posts, and literary text datasets in Hiligaynon using Transformer architectures mBERT and XLM-RoBERTa.}
    \label{fig:1}
\end{figure*}

\section{Building \textsc{HiligayNER}: A Baseline NER Model for Hiligaynon}
\subsection{Dataset Collection}
\textsc{HiligayNER} was assembled in three sequential phases: data collection, expert annotation, and reliability testing. Five online platforms hosting publicly available content were crawled to capture a sizeable representation of contemporary Hiligaynon texts as reported in Table~\ref{table:1}. Each row in Table 1 refers to a single sentence extracted from the respective source. The dataset was segmented at the sentence level to facilitate BIO tagging and sentence-level NER annotations. The initial, raw collection comprised 17,647 sentences, but was reduced to 8,082 after preprocessing to remove malformed strings, empty lines, and non-Hiligaynon texts.

\begin{table}[!htbp]
\centering
\resizebox{\columnwidth}{!}{%
\begin{tabular}{lrr}
\toprule
\textbf{Source} & \textbf{Original} & \textbf{Cleaned} \\
\midrule
Ang Pulong Sang Dios & 11{,}000 & 5{,}500 \\
Ilonggo News Live & 3{,}925 & 1{,}877 \\
Hiligaynon News and Features & 2{,}281 & 276 \\
Bombo Radyo Bacolod & 286 & 276 \\
Ilonggo Balita sa Uma & 155 & 153 \\
\bottomrule
\end{tabular}%
}
\caption{Statistics of publicly available data sources used in building \textsc{HiligayNER}.}
\label{table:1}
\end{table}

\subsection{Annotation Process and Reliability Testing}
Three (3) undergraduate linguistics students who are also native speakers of Hiligaynon were tasked to annotate the corpus using Label Studio \cite{r21}. The guidelines for annotating follow the CoNLL-2003 BIO convention \cite{tjong-kim-sang-de-meulder-2003-introduction} with four entity categories: Person (B-PER and I-PER), Organization (B-ORG and I-ORG), Location (B-LOC and I-LOC), and Other (OTH). For reference, in BIO tagging for NER, the B-prefix represents the first token of a named entity, while the I-prefix represents subsequent terms of a named entity. Refer to an example of a tagged sentence below using the BIO convention: \\\\
\noindent
\fcolorbox{blue}{white}{\textbf{B-PER} Aling} \quad
\fcolorbox{blue}{white}{\textbf{I-PER} Myrna} \quad
\fcolorbox{black}{white}{\textbf{O} went} \quad
\fcolorbox{black}{white}{\textbf{O} to} \quad
\fcolorbox{red}{white}{\textbf{B-LOC} Iloilo} \quad
\fcolorbox{red}{white}{\textbf{I-LOC} City}.\\

The annotators received ten hours of joint training, including pilot rounds on 250 sentences with adjudication by a supervising linguist. Disagreements were resolved through consensus meetings and the final labels were exported in CoNLL format. To assess the reliability of the annotations, a stratified 10\% subset of the corpus was selected and annotated independently by all three annotators. Cohen’s $\kappa$ was then computed based on pairwise comparisons within this overlapping subset to measure annotation consistency. Cohen’s $\kappa$, a statistical metric widely adopted in NER studies \cite{r24,r25}. The remaining portion of the dataset was divided among annotators for individual annotation. Table ~\ref{table:2} reports on the scores showing an observed agreement = 0.9493, expected agreement = 0.7273, and yielding $\kappa = 0.8141$. According to conventional interpretation, a $\kappa \geq$ 0.80 equates to substantial agreement, which denotes that the annotations of the \textsc{HiligayNER} dataset are of high quality and suitable for reproducible model training.

\begin{table}[!htbp]
\centering
\begin{tabular}{lc}
\toprule
\textbf{Metric} & \textbf{Value} \\
\midrule
Observed Agreement & 0.9493 \\
Agreement by Chance & 0.7273 \\
{Cohen's $\kappa$} & {0.8141} \\
\bottomrule
\end{tabular}
\caption{Cohen’s $\kappa$ agreement results from annotations.}
\label{table:2}
\end{table}

\subsection{Finetuning}
To establish strong baselines for \textsc{HiligayNER}, we fine-tuned two multilingual transformer encoders Multilingual BERT (mBERT) and XLM-RoBERTa (XLM-R) using the standard token-classification pipeline in Hugging Face Transformers \cite{r27}. Both models are pretrained on large cross-lingual corpora and have shown competitive zero-shot and few-shot performance on sequence-labelling tasks \cite{r29,r30}.

\paragraph{Multilingual BERT (mBERT).}
mBERT is a 12-layer, 768-hidden, 12-head encoder trained on Wikipedia dumps from 104 languages \cite{devlin-etal-2019-bert}. For NER, we attach a softmax-classifier head that maps each contextual token representation ht to a probability distribution over the four entity tags (PER, ORG, LOC, OTH):

\begin{equation}
P(\hat{y}|x=\prod_{t=1}^{T}softmax(Wh_t+b)    
\end{equation}

\paragraph{XLM-RoBERTa.}
XLM-RoBERTa extends the vanilla RoBERTa architecture \cite{r28} to 100 languages, pretrained on 2.5 TB of CommonCrawl with a SentencePiece tokenizer and larger capacity (24 layers, 1024 hidden, 16 heads) \cite{r29}. We replicated the mBERT fine-tuning recipe but lowered the learning rate to 3×$10^{-5}$, following XLM-R recommendations. Empirically, XLM-R attains higher recall on low-frequency tags, confirming earlier cross-lingual findings \cite{r29}.

\section{Result and Discussion}

\subsection{Training mBERT and XLM-RoBERTa}
\begin{figure}[htbp]
    \centering
    \includegraphics[width=0.45\textwidth]{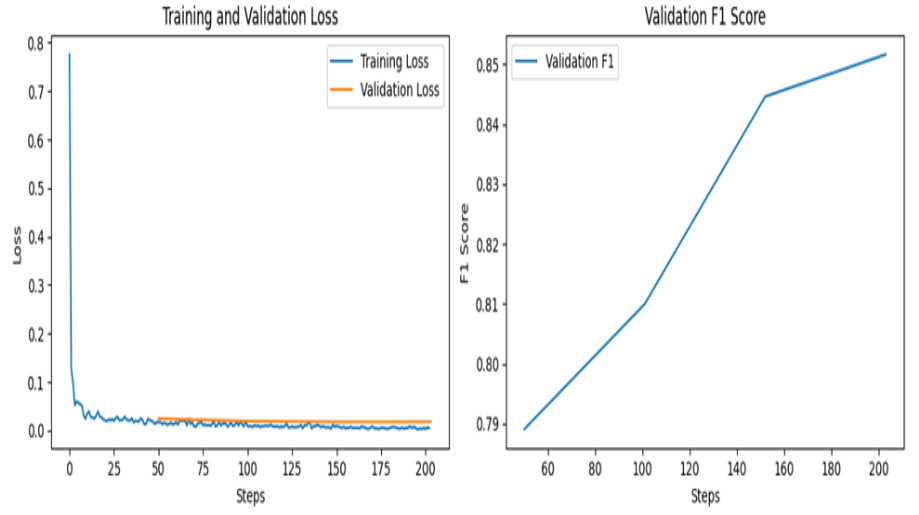}
    \caption{Training loss, validation loss, and F1 score per training step for the finetuned mBERT model.}
    \label{fig:4}
\end{figure}

\begin{figure}[htbp]
    \centering
    \includegraphics[width=0.45\textwidth]{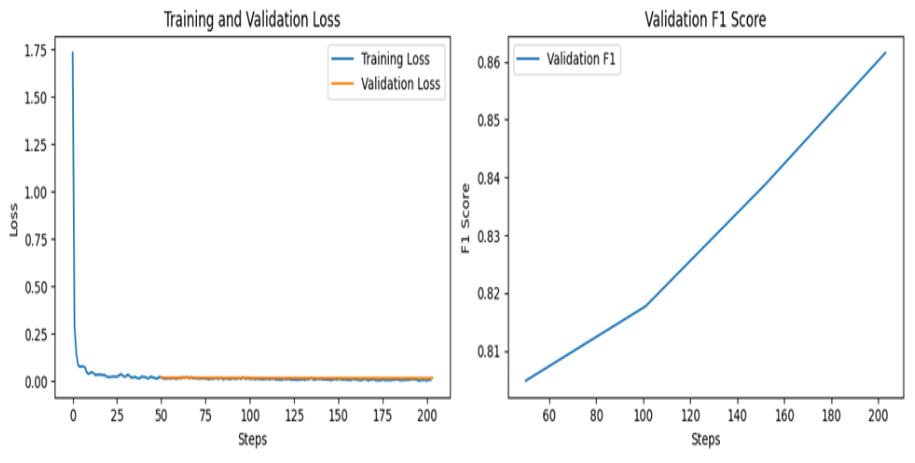}
    \caption{Training loss, validation loss, and F1 score per training step for the finetuned XLM-RoBERTa model.}
    \label{fig:5}
\end{figure}

Figures ~\ref{fig:4} and ~\ref{fig:5} plot the optimization trajectories for mBERT and XLM-RoBERTa, respectively. In both cases, the training loss decays monotonically during the first 100 batches and flattens thereafter, signaling rapid convergence under the chosen hyperparameters. Validation loss closely tracks the training curve and stabilizes at <0.05, indicating an absence of over-fitting.


Figures ~\ref{fig:4} and ~\ref{fig:5} reveal rapid, stable convergence. Training loss drops sharply and levels off; validation loss mirrors this trajectory, remaining below 0.05. F1 improves in tandem—mBERT from 0.79 to 0.87, XLM-R to 0.88—without divergence between training and validation curves. The results confirm that the three-epoch, AdamW fine-tuning regimen achieves generalisation without over-fitting.

\subsection{Model Evaluation}
Tables ~\ref{table:3} and ~\ref{table:4} report token-level precision, recall, and F1 for the two Transformer-based models. In the case of mBERT, the model attains a macro F1 of 0.86, with near-perfect recognition of Person-based named entities at 0.96 and 0.94 for B-PER and I-PER. Location-based entities follow as the second-most correctly recognized at 0.83 and 0.82 for B-LOC and I-LOC. At the same time, Organization remains the most challenging entity to recognize for mBERT at 0.82 and 0.79. Nonetheless, these values are all relatively decent performances given that they exceed the 0.80 benchmark. 

In the case of XLM-RoBERTa, we see a comparable high performance where Person-based entities are the most correctly recognized span, giving 0.96 and 0.94 for B-PER and I-PER. Location entities scored moderately, with B-LOC of 0.82 and I-LOC of 0.84 for F1, while organization entities remained the most challenging, yielding 0.81 for B-ORG and 0.79 for I-ORG.

For both models, we observe a general pattern where performance metrics correlate with entity tag frequency, with higher scores in categories with larger support counts (e.g., I-PER with 2,181 instances) compared to less frequent categories such as B-ORG (505 cases). These findings are consistent with prior multilingual-NER evaluations showing that pretrained transformers handle person names best and struggle with organization boundary cues \cite{r29,r31}.

The study reports token-level precision, recall, and F1 scores as the primary evaluation metrics. Entity-level evaluation was not conducted, as the scope of this work is to establish a baseline for Hiligaynon NER using token-level annotation and modeling. The evaluation approach follows the convention used in the recently published CebuaNER study  \cite{pilar-etal-2023-cebuaner}, which also adopted token-level reporting as a standard for establishing baselines in low-resource Philippine languages. The researchers recognize that span-level evaluation provides a stricter measure of system performance and leave this as an important direction for future work.

\begin{table}[t]
\centering
\resizebox{\columnwidth}{!}{%
\begin{tabular}{lrrrr}
\toprule
\textbf{Tagset} & \textbf{Precision} & \textbf{Recall} & \textbf{F1-Score} & \textbf{Support} \\
\midrule
B-PER & 0.95 & 0.97 & 0.96 & 1{,}754 \\
I-PER & 0.93 & 0.94 & 0.94 & 2{,}181 \\
B-LOC & 0.79 & 0.86 & 0.83 & 565 \\
I-LOC & 0.82 & 0.83 & 0.82 & 1{,}237 \\
B-ORG & 0.77 & 0.87 & 0.82 & 505 \\
I-ORG & 0.77 & 0.82 & 0.79 & 944 \\
\bottomrule
\end{tabular}%
}
\caption{Performance of the finetuned mBERT model using \textsc{HiligayNER} across NER categories.}
\label{table:3}
\end{table}

\begin{table}[t]
\centering
\resizebox{\columnwidth}{!}{%
\begin{tabular}{lrrrr}
\toprule
\textbf{Tagset} & \textbf{Precision} & \textbf{Recall} & \textbf{F1-Score} & \textbf{Support} \\
\midrule
B-PER & 0.95 & 0.97 & 0.96 & 1{,}777 \\
I-PER & 0.93 & 0.95 & 0.94 & 2{,}268 \\
B-LOC & 0.79 & 0.86 & 0.82 & 577 \\
I-LOC & 0.83 & 0.85 & 0.84 & 1{,}228 \\
B-ORG & 0.76 & 0.87 & 0.81 & 514 \\
I-ORG & 0.74 & 0.84 & 0.79 & 910 \\
\bottomrule
\end{tabular}%
}
\caption{Performance of the finetuned XLM-RoBERTa model using \textsc{HiligayNER} across NER categories.}
\label{table:4}
\end{table}

\subsection{Error Analysis}
\begin{figure}[t]
    \centering
    \includegraphics[width=0.45\textwidth]{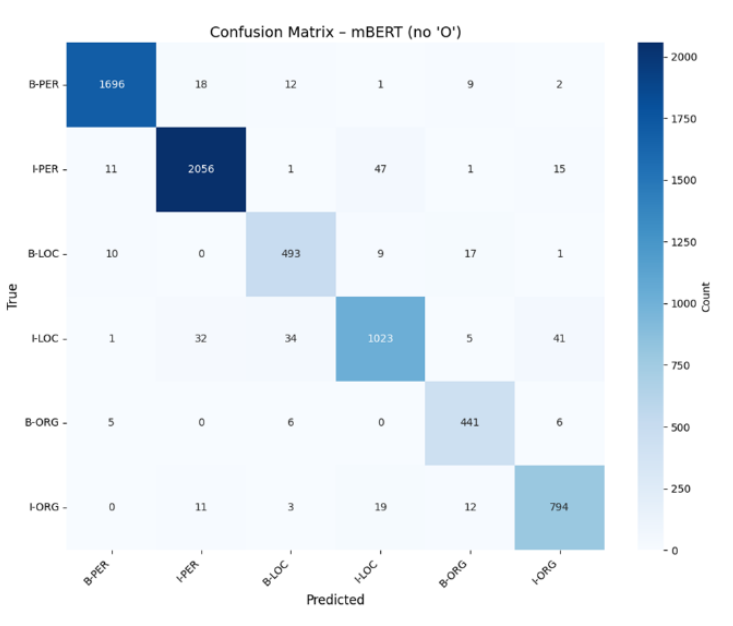}
    \caption{Confusion matrix of the finetuned mBERT model using \textsc{HiligayNER} across NER categories, omitting the OTH tag for brevity.}
    \label{fig:2}
\end{figure}

\begin{figure}[t]
    \centering
    \includegraphics[width=0.45\textwidth]{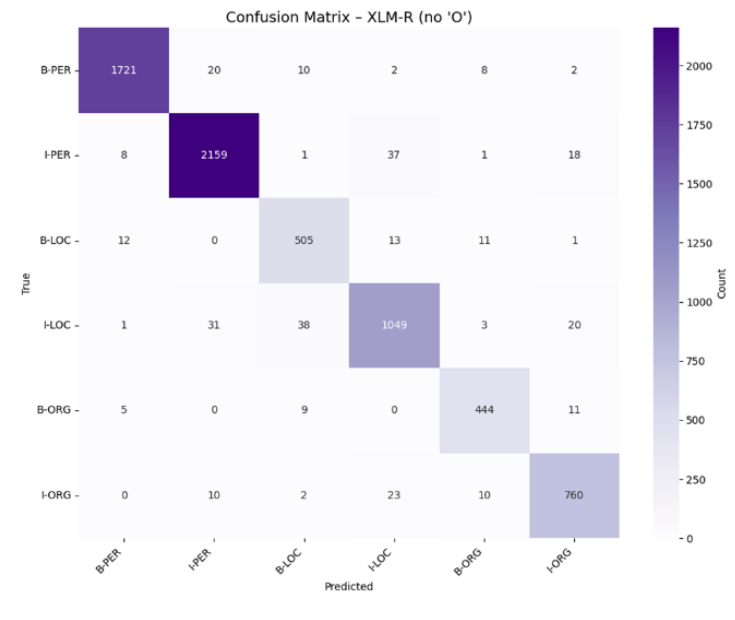}
    \caption{Confusion matrix of the finetuned XLM-RoBERTa model using \textsc{HiligayNER} across NER categories, omitting the OTH tag for brevity.}
    \label{fig:3}
\end{figure}

Figures ~\ref{fig:2} and ~\ref{fig:3} expose the distribution of residual errors after fine-tuning for XLM-RoBERTa and mBERT, respectively. In both matrices, person entities dominate the main diagonal B-PER and I-PER account for > 96\% of their respective instances, confirming that multilingual transformers consistently capture personal-name cues. Both models maintain negligible cross-category bleed between person and non-person tags (< 0.5\%), and false positives for rare classes remain below 1\% of total predictions. The matrices, therefore, corroborate the aggregate metrics where the entity segmentation is reliable for PER, adequate for LOC, and bottlenecked by ORG boundary precision. Targeted gazetteer augmentation or span-level objectives should prioritize the ORG–LOC boundary to yield substantive gains.

\subsection{Crosslingual Performance with Cebuano and Tagalog}
Table ~\ref{table:5} presents the crosslingual performance of both the mBERT and XLM-RoBERTa models finetuned on \textsc{HiligayNER}. Results from zero-shot evaluation on Cebuano and Tagalog yield macro F1 scores of $\approx$ 0.46 (0.44 to 0.46) for both languages, which are comparable to earlier Philippine cross-lingual results \cite{r11,r28}. Precision, on the other hand, is marginally higher on Cebuano, reflecting closer lexical affinity within the Central Philippine subgroup \cite{imperial-kochmar-2023-automatic,imperial-kochmar-2023-basahacorpus}. Although lower than in-language scores, the outcome demonstrates that the released model checkpoints offer a viable starting point for rapid adaptation to neighboring languages. The higher performance of Cebuano over Tagalog may be attributed to its lexical and syntactic proximity to Hiligaynon, as both belong to the Central Philippine language subgroup and share similar morphological patterns and word order. In contrast, Tagalog, while still within the same Austronesian family, exhibits more divergent lexical structures. It is also worth mentioning that Cebuano, Tagalog, and Hiligaynon are written using the Latin script, which may have contributed to their crosslingual generalization.


\begin{table}[htbp]
\centering
\resizebox{\columnwidth}{!}{%
\begin{tabular}{lcccc}
\toprule
\textbf{Metrics} & \multicolumn{2}{c}{\textbf{mBERT}} & \multicolumn{2}{c}{\textbf{XLM-RoBERTa}} \\
\cmidrule(lr){2-3} \cmidrule(lr){4-5}
{} & \textbf{CEB} & \textbf{TAG} & \textbf{CEB} & \textbf{TAG} \\
\midrule
Precision & 0.4402 & 0.3998 & 0.4340 & 0.3894 \\
Recall    & 0.4773 & 0.4991 & 0.4984 & 0.5221 \\
F1-Score  & 0.4580 & 0.4439 & 0.4640 & 0.4461 \\
Accuracy  & 0.9727 & 0.9639 & 0.9736 & 0.9633 \\
\bottomrule
\end{tabular}%
}
\caption{Cross-lingual performance of the finetuned mBERT and XLM-RoBERTa models using \textsc{HiligayNER} with Cebuano and Tagalog languages.}
\label{table:5}
\end{table}



\section{Conclusion}
This study presents \textsc{HiligayNER}, the first publicly available baseline NER model and dataset for Hiligaynon, a digitally underrepresented regional language in Western Visayas, Philippines. The \textsc{HiligayNER} dataset was systematically annotated under CoNLL BIO guidelines by native speakers and validated with strong inter-annotator agreement ($\kappa = 0.81$). Finetuning experiments on two multilingual models {mBERT} and {XLM-RoBERTa} yielded macro F1 $\approx$ 0.86, surpassing the 0.80 threshold on all primary tags (Person, Location, and Organization), which presents a high-quality baseline performance. Additional error analysis showed that residual confusion is concentrated in organization–location boundaries, while zero-shot transfer to Cebuano and Tagalog achieved competitive F1 $\approx$ 0.46, confirming cross-lingual utility. 

By releasing the corpus, annotation protocol, training scripts, and model checkpoints under a permissive license, we provide a reproducible foundation for downstream Hiligaynon NLP and rapid adaptation to related Central Philippine languages \cite{imperial-kochmar-2023-automatic,imperial-kochmar-2023-basahacorpus}. For future work, we recommend further efforts on increasing and diversifying the content of \textsc{HiligayNER}, such as adding finer-grained tags (e.g., Event, Date), exploring domain-adaptive pre-training on regional news, and incorporating gazetteer-augmented span objectives to improve organization recognition. These directions will further advance language technology for Hiligaynon and other low-resource languages.

\section*{Acknowledgments}
All datasets collected for this study are publicly available and are used for non-commercial research purposes. We acknowledge the sources of the Hiligaynon data from Ang Pulong Sang Dios, Ilonggo News Live, Hiligaynon News and Features, Bombo Radyo Bacolod, and Ilonggo Balita sa Uma. We gratefully acknowledge the financial support provided by the National University and the Department of Science and Technology for the General Access Multilingual Online Tool for Public Health Drug-Reporting (GamotPH) Project. 

\bibliography{custom,anthology}



\end{document}